\documentclass[10pt,twocolumn,letterpaper]{article}

\usepackage{cvpr}
\usepackage{times}
\usepackage{epsfig}
\usepackage{graphicx}
\usepackage{array,multirow}
\usepackage{amsmath}
\usepackage{amssymb}
\usepackage{booktabs}
\usepackage{algorithmicx}

\usepackage[pagebackref=true,breaklinks=true,letterpaper=true,colorlinks,bookmarks=false]{hyperref}

\usepackage{color}
\usepackage[ruled,linesnumbered]{algorithm2e}

\graphicspath{{./img/}}

\cvprfinalcopy 

\def\cvprPaperID{6127} 

\ifcvprfinal\fi
\begin{document}

\title{HyperSTAR: Task-Aware Hyperparameters for Deep Networks}

\author{Gaurav Mittal$^\star$$^\dag$ ~~~~ Chang Liu$^\star$$^\ddag$ ~~~~ Nikolaos Karianakis$^\dag$ ~~~~ Victor Fragoso$^\dag$ ~~~~  Mei Chen$^\dag$ ~~~~ Yun Fu$^\ddag$\\
$^\dagger$Microsoft ~~~~~~~~~~~~~~~~~~~~~~~~ $^\ddag$Northeastern University\\
{\tt\small \{gaurav.mittal, nikolaos.karianakis, victor.fragoso, mei.chen\}@microsoft.com} \\
{\tt\small liu.chang6@husky.neu.edu  ~~~~ yunfu@ece.neu.edu}
}

\maketitle
\thispagestyle{empty}


\begin{abstract}
While deep neural networks excel in solving visual recognition tasks, they require significant effort to find hyperparameters that make them work optimally. Hyperparameter Optimization (HPO) approaches have automated the process of finding good hyperparameters but they do not adapt to a given task~(task-agnostic), making them computationally inefficient. To reduce HPO time, we present HyperSTAR (System for Task Aware Hyperparameter Recommendation), a task-aware method to warm-start HPO for deep neural networks. HyperSTAR ranks and recommends hyperparameters by predicting their performance conditioned on a joint dataset-hyperparameter space. It learns a dataset~(task) representation along with the performance predictor directly from raw images in an end-to-end fashion. The recommendations, when integrated with an existing HPO method, make it task-aware and significantly reduce the time to achieve optimal performance. We conduct extensive experiments on 10 publicly available large-scale image classification datasets over two different network architectures, validating that HyperSTAR evaluates 50\% less configurations to achieve the best performance compared to existing methods. We further demonstrate that HyperSTAR makes Hyperband (HB) task-aware, 
achieving the optimal accuracy in just $25\%$ of the budget required by both vanilla HB and Bayesian Optimized HB~(BOHB).
\end{abstract}


\vspace{-3mm}
\section{Introduction}
\vspace{-3mm}

\let\thefootnote\relax\footnote{$^\star$ Authors with equal contribution. Published at CVPR 2020 (Oral).}
\let\thefootnote\relax\footnote{This work was done when C. Liu was a research intern at Microsoft.}

Transfer learning has become a de-facto practice to push the performance boundary on several computer vision tasks~\cite{donahue2014decaf, yosinski2014transferable}, most notably image classification~\cite{he2016deep, huang2017densely, Simonyan2014VeryDC}. 
Although transfer learning improves performance on new tasks, it requires machine learning (ML) experts to spend hours finding the right hyperparameters (\eg, learning rate, layers to fine-tune, optimizer, \etc) that can achieve the best performance. Researchers have relied on Hyperparameter Optimization (HPO), ranging from simple random search~\cite{bergstra2012random} to sophisticated Bayesian Optimization~\cite{snoek2012practical} and Hyperband~\cite{li2016hyperband}, to reduce manual effort and automate the process of finding the optimal hyperparameters. Though more effective than manual search, these approaches are still slow since most of them trigger the same procedure for any new task and do not leverage any information from past experiences on related tasks.

\begin{figure}[t]
    \label{fig:teaser}
    \centering
    \includegraphics[width=\columnwidth]{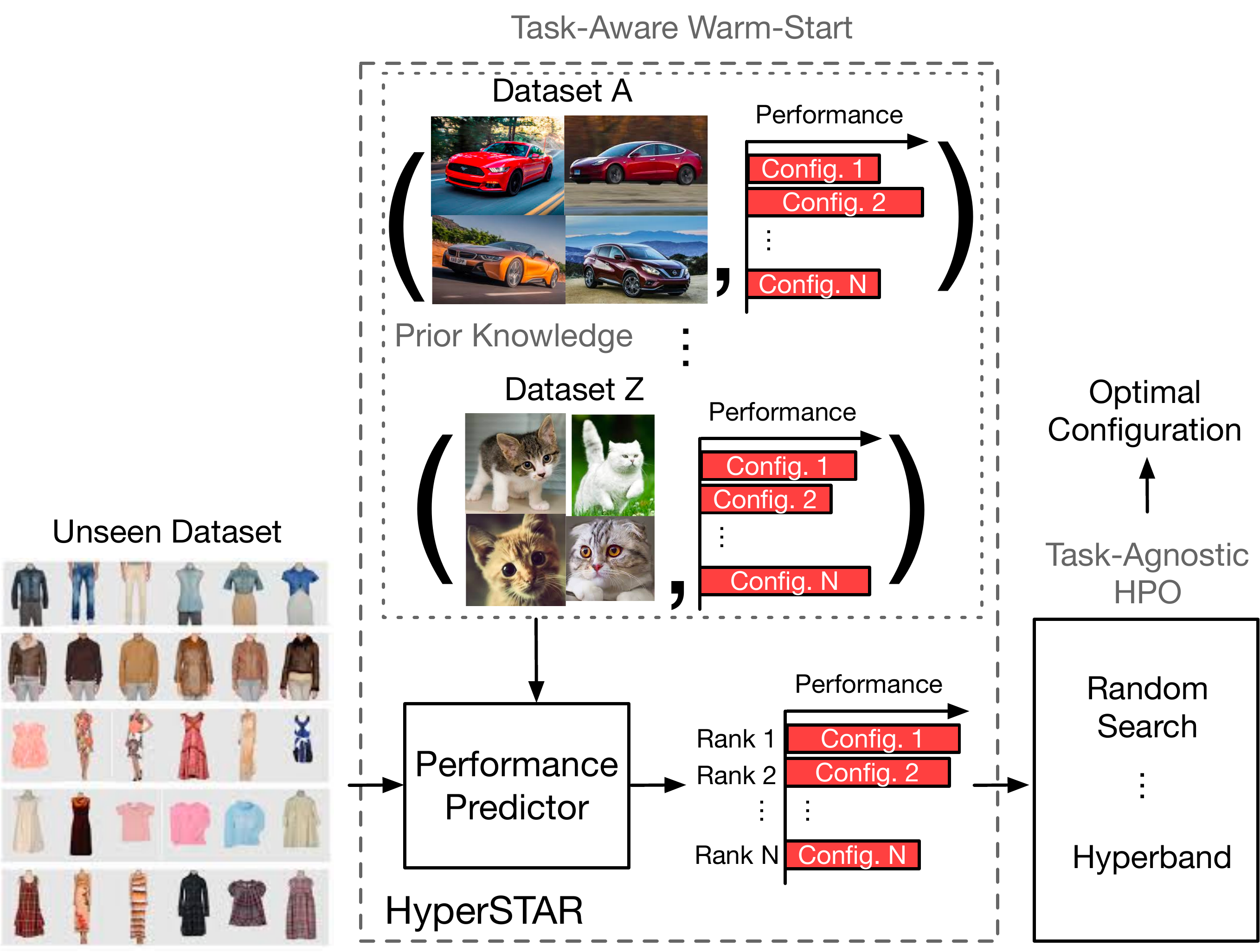}
    \caption{HyperSTAR learns to recommend optimal hyperparameter configurations for an unseen task by learning end-to-end over a joint dataset-hyperparameter space. These recommendations can accelerate existing HPO methods leading to state-of-the-art performance for resource-constrained budgets.}
    \vspace{-10pt}
\end{figure}

Many approaches accelerate HPO~\cite{kandasamy2017multi, klein2016learning, swersky2013multi} including ``warm-start'' techniques which exploit the correlation among tasks~\cite{swersky2013multi,perrone2018scalable, feurer2015initializing, bardenet2013collaborative, xue2019transferable}. Some of these use meta-learning to warm-start HPO by exploiting the task information~(meta-features) from past searches~\cite{feurer2015initializing}. These methods either guide the search policy for hyperparameters using a learned prior over hand-crafted dataset statistics~\cite{feurer2015initializing, yogatama2014efficient, bardenet2013collaborative}, or pick the search policy of the most similar task from a database~\cite{xue2019transferable, feurer2015efficient}. Although these methods accelerate HPO, there is no method that leverages visual-based priors or learns deep feature representations to jointly encode dataset and hyperparameters to expedite HPO on large-scale image classification tasks. Having such a representation can help systems warm-start and tailor an HPO method based on the task to optimize. While Kim~\etal~\cite{kim2017learning} and Wong~\etal~\cite{wong2018transfer} suggest using image features to understand the task, their efforts lack a joint representation for the tasks and hyperparameters. We argue that a joint dataset-hyperparameter representation is crucial for large-scale, real-world image classification problems.

With the advent of AutoML~\cite{automl}, there is a strong interest for systems~\cite{NIPS2015_5872, jin2019auto} to fully automate the process of training a model on a customer image dataset. To cater to a large number of users, it is essential for AutoML systems to be efficient in searching for the optimal hyperparameters. Given the diversity of real-world image datasets, it is also necessary to prioritize the hyperparameter configurations in a task-aware manner rather than being task-agnostic. A task-aware mechanism understands a given dataset and recommends configurations that can operate well on that dataset. On the other hand, a task-agnostic mechanism treats all datasets equally and sets off the same configuration search regardless of the task.



In order to enable task-aware HPO, we introduce HyperSTAR (System for Task-Aware Recommendation), a warm-start algorithm that prioritizes optimal hyperparameter configurations for an unseen image classification problem. HyperSTAR learns to recommend hyperparameter configurations for a new task from a set of previously-seen datasets and their normalized performance over a set of hyperparameter configurations. It comprises of two phases: an offline meta-learning phase and an online recommendation phase. In the meta-learning phase, HyperSTAR trains a network
to first learn a task representation for a given dataset directly from its training images. Then, it uses the representation to learn an accuracy predictor for a given configuration. In the recommendation phase, HyperSTAR predicts the accuracy for each hyperparameter configuration given the task representation of an unseen dataset. It then exploits these predictions to generate a ranking which can be used to accelerate different HPO approaches by prioritizing the most promising configurations for evaluation. 
See Fig.~\ref{fig:teaser} for an illustration of HyperSTAR.



Our extensive ablation studies demonstrate the effectiveness of HyperSTAR in recommending configurations for real-world image classification tasks. We also formulate a task-aware variant of Hyperband (HB)~\cite{li2016hyperband} using the recommendation from HyperSTAR and show that it outperforms previous variations~\cite{li2016hyperband, falkner18abohb, xue2019transferable} in limited time budget HPO settings. 
To the best of our knowledge, HyperSTAR is the first warm-starting method that learns to accelerate HPO for large-scale image classification problems from hyperparameters and raw images in an end-to-end fashion.

In sum, the contributions of this work are the following:
\begin{itemize}
    \item A meta-learning framework, HyperSTAR, that recommends task-specific optimal hyperparameters for unseen real-world image datasets.
    \item The first method to recommend hyperparameters based on a task-representation learned jointly with a performance predictor end-to-end from raw images.
    \item HyperSTAR can warm-start and accelerate task-agnostic HPO approaches. We demonstrate this by integrating HyperSTAR with Hyperband which outperforms existing methods in limited budget setting.  
\end{itemize}
\section{Related Work}
\begin{figure*}[h]
    \label{fig:overview}
    \centering
    \includegraphics[width=\textwidth]{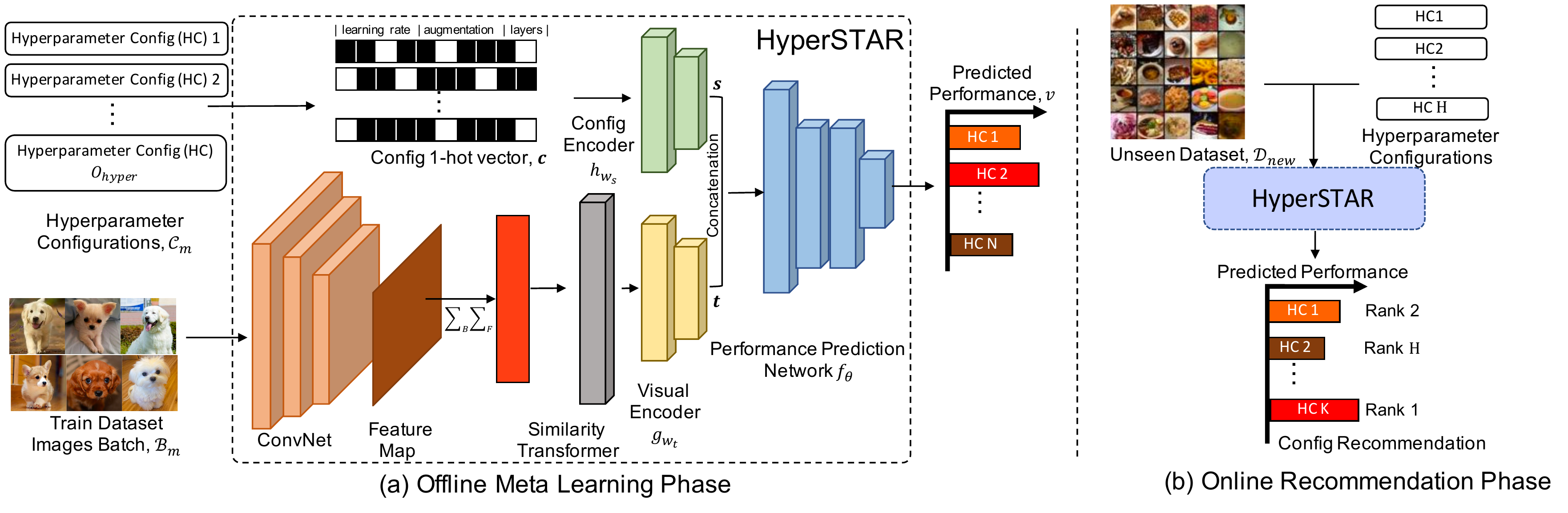}
    \caption{HyperSTAR Model Overview. (a) \textbf{Offline Meta-Learning Phase}. This phase jointly learns the functions for task representation and hyperparameter representation, and uses them as input to a performance predictor that estimates the performance of a CNN given a dataset (task) and a hyperparameter configuration. (b) \textbf{Online Recommendation Phase}. In this phase, HyperSTAR predicts the performance over the hyperparameter space for a new dataset and generates a task-aware ranking of the configurations.}
    \vspace{-4mm}
\end{figure*}

The simplest solution to find hyperparameters for an algorithm is via a grid search over all the possible parameters~\cite{bergstra2012random}. Since it is slow and computationally expensive, the community introduced methods such as Bayesian Optimization~(BO)~\cite{snoek2012practical, snoek2015scalable, klein2017fast} that use Gaussian processes for probabilistic sampling and Hyperband~\cite{li2016hyperband} which uses random configuration selection and successive halving~\cite{jamieson2016non} to speed up HPO. Falkner~\etal~\cite{falkner18abohb} proposed BOHB, a Bayesian optimization and Hyperband hybrid that exploits the tradeoff between performance and time between BO and HB. For low time budgets, BOHB and Hyperband are equally better than BO while for large time budgets, BOHB outperforms all BO, Hyperband, and random search~\cite{automl}.

To accelerate HPO approaches, there are methods that model learning curves~\cite{swersky2014freeze,klein2016learning}, use multi-fidelity methods for cheap approximations~\cite{kandasamy2017multi}, use gradient-based methods\cite{franceschi2017forward, maclaurin2015gradient, PedregosaHyperparameter16}, or train on a subset of training data and extrapolate the performance~\cite{klein2017fast} to reduce the overall search time. An alternative way to speed up HPO is via ``warm-start'' techniques~\cite{swersky2013multi,perrone2018scalable, feurer2015initializing, bardenet2013collaborative, xue2019transferable}. These techniques exploit correlation between tasks to accelerate HPO. Swersky~\etal~\cite{swersky2013multi} learns to sample hyperparameters based on multi-task Gaussian processes. Xue~\etal\cite{xue2019transferable} clusters previously-evaluated tasks based on the accuracy on certain benchmark models. Both approaches, while exploiting HPO knowledge from multiple tasks, incur a time overhead as they need to evaluate the new task every time over a certain pool of configurations in order to speed up the search.

To avoid evaluating benchmark configurations, other approaches learn a function to map the trend in performance of a task over the configuration space with some task-based representation~\cite{bardenet2013collaborative, feurer2015initializing, lindauer2018warmstarting, yogatama2014efficient, feurer2015efficient}. This function is based on multi-task Gaussian processes~\cite{bardenet2013collaborative, yogatama2014efficient} or random forests~\cite{feurer2015initializing}. The task representations employed in these methods are based on hand-crafted features such as meta data (\eg, number of samples and labels in the dataset), or first and second order statistics (\eg, PCA, skewness, kurtosis, \etc)~\cite{automl}. Since these features are neither visual-based nor learned jointly with the HPO module, they prove to be inefficient for large-scale vision tasks (see Section~\ref{experiments}).

Achille~\etal~\cite{achille2019task2vec} introduced task2vec, a visual-inspired task representation but its computational cost makes it ill-suited as conditional input for hyperparameter recommendation. With respect to neural architectures~\cite{zoph2016neural}, Kokiopoulou~\etal ~\cite{kokiopoulou2019fast} suggests conditioning the architecture search for natural language tasks over globally averaged features obtained from raw language based data. However, being restricted to low dimensional language tasks and without any dedicated performance based similarity regularization, these methods are not directly applicable to and effective on large scale vision tasks. For vision tasks, Wong~\etal and Kim~\etal \cite{wong2018transfer, kim2017learning} condition the search of architectures and/or hyperparameters over deep visual features globally averaged over all images. Unlike these methods where features are either not learned or are aggregated via simple statistics (\ie, a global mean), HyperSTAR is the first method that learns an end-to-end representation over a joint space of hyperparameters and datasets. By doing so, HyperSTAR learns features that are more actionable for recommending configurations and for task-aware warm-start of HPO for large-scale vision datasets.

\section{HyperSTAR}
 The goal of HyperSTAR is to recommend tailored hyperparameter configurations for an unseen dataset (task). To introduce this task-awareness, HyperSTAR comprises of a supervised performance predictor operating over a joint space of real-world image classification datasets and hyperparameter configurations. Given a dataset and a hyperparameter configuration, our model learns to predict the performance of the dataset for the given configuration in an offline meta-learning phase. Once the model has learned this mapping, we use HyperSTAR on an unseen dataset in an online recommendation phase to predict scores and rank the hyperparameter configurations. This ranking is beneficial to warm start task-agnostic HPO approaches, as we demonstrate via our formulation of task-aware Hyperband. Figure~\ref{fig:overview} provides a detailed illustration of HyperSTAR.
\vspace{-1mm}
\subsection{Offline Meta-Learning Phase}
\vspace{-1mm}
\label{sec:offline_meta}
{\noindent \bf Performance Predictor.}
The objective of the performance predictor is to estimate the accuracy of a hyperparameter configuration given a task representation and an encoding of the hyperparameters (\eg, learning rate, number of layers to fine-tune, optimizer). Mathematically, the performance predictor $f$ is a function that regresses the performance $v$ of a deep-learning based image classifier given a dataset or task $\mathcal{D}$ and a hyperparameter configuration encoding $\mathcal{C}$.

Because deriving this function $f$ analytically is challenging for real-world vision tasks, we instead learn it using a deep network architecture $f_\theta$ parameterized with weights $\theta$. Learning $f_\theta$ requires a tensor-based representation of the dataset $\mathcal{D}$ and hyperparameter configuration $\mathcal{C}$. To learn the representation of the task $\mathbf{t} \in \mathbb{R}^d$, we search for a function $\mathbf{t} = g_{w_t}(\mathcal{D})$ parameterized by weights $w_t$. Similarly, we learn the representation $\mathbf{s} \in \mathbb{R}^d$ of a hyperparameter configuration encoded as one-hot vector $\mathbf{c}$ by searching for a function $\mathbf{s} = h_{w_s}(\mathbf{c})$ parameterized by the weights $w_s$.  Mathematically, this is formulated as $v = f_\theta \left(g_{w_t}\left( \mathcal{D}\right), h_{w_s}(\mathbf{c})\right)$.

We learn the task representation $\mathbf{t}$ in an end-to-end manner directly from raw training images of dataset $\mathcal{D}$ by using a convolutional neural network followed by a transformer layer inside $g_{w_t}(\cdot)$. This enables the use of visual information into the task representation $\mathbf{t}$, leading to improved generalization over unseen vision tasks and making the method end-to-end differentiable. 
Jointly learning the performance predictor and the representations in an end-to-end fashion constitutes a departure from previous meta-learning approaches that represent a task using hand-crafted metadata~\cite{bardenet2013collaborative} (\eg, total number of training samples, number of classes, number of samples per class, \etc), performance-based features~\cite{xue2019transferable} or globally averaged features from a frozen deep network~\cite{wong2018transfer}. This allows our performance predictor $f_\theta \left( \cdot \right)$ to inform the feature extractors $g_{w_t}(\cdot)$ and $h_{w_s}(\cdot)$ during training about the most useful features for estimating the performance of an image classifier  given a task and a hyperparameter configuration.

{\noindent \bf Meta Dataset.} To jointly learn the performance predictor $f_{\theta}\left(\cdot\right)$, task representation $g_{w_t}(\cdot)$, and hyperparameter embedding $h_{w_s}(\cdot)$ in a supervised manner, we construct a meta dataset (\ie, a dataset of datasets) $\mathcal{T}$ over the joint space of $M$ datasets and $H$ hyperparameter configurations. We define $\mathcal{T} =   \left\{ (\mathcal{D}_{i}, \mathbf{c}_j ,  v_{ij}) \; | \; i \in \{1, \ldots, M\}, j \in \{1, \ldots, H\}  \right  \}$, 
where $v_{ij}$ is the target performance score (\eg, top-1 accuracy) achieved for an image classifier using the hyperparameter configuration $\mathbf{c}_j$ on a dataset $\mathcal{D}_{i} =   \{ (x^{k}_{i}, y^{k}_i ) \; | \;  k  = 0, \ldots, N_{i} \}$ (each ($x^{k}_{i}, y^k_i$) being an image-label pair in $\mathcal{D}_i$).

{\noindent \bf  Performance Regression.} We first find the optimal parameters for $\theta, w_t, w_s$ that minimize the difference between the estimated performance of our performance predictor and the ground-truth performance score using the loss function:
\begin{equation}
    \mathcal{L}_{\text{perf}}(w_t,w_s,\theta) = \frac{1}{B}\sum_{i=1}^B \left\| v_{ij} - f_\theta \left(g_{w_t}\left(\mathcal{D}_i \right), h_{w_s}(\mathbf{c}_j)\right) \right\|_2^2,
    \label{eq:loss_fn}
\end{equation}
where $B$ is the number of instances in a batch.

The raw ground truth performance scores $v_{ij}$ across datasets can have a large variance due to the diversity of task difficulty. To alleviate the effect of this variance on our predictor, we normalize the performance scores as, $v_{ij} \leftarrow (v_{ij} - \mu_{i})\sigma^{-1}_{i}$, where $\mu_i$ and $\sigma_i$ are the mean and standard deviation over the performance scores of dataset $\mathcal{D}_i$ for all hyperparameter configurations, respectively.

Although formulating the objective function using a ranking loss~\cite{chen2009ranking} seems more intuitive for recommending hyperparameters, Yogatama~\etal~\cite{yogatama2014efficient} showed that applying the above normalization over $v_{ij}$ makes the regression-based optimization in Eq.~\eqref{eq:loss_fn} equivalent to a rank-based optimization. A regression-based formulation has the advantage of being more time-efficient than rank-based optimization. The time complexity of learning a rank-based predictor is $\mathcal{O}(M H^{2} )$ while that of a regression-based predictor is $\mathcal{O}(MH)$. Consequently, our regression-based performance predictor can scale favorably to many more datasets.

{\noindent \bf  Similarity-based inter-task regularization.}
To learn a more meaningful task representation, we add a regularizer that imposes that two tasks must have similar representations if they have similar hyperparameter-configuration rankings. The goal of this regularizer, $\mathcal{L}_{\text{sim}}(w_t)$, is to penalize our model when the similarity of two task representations differs from a pre-computed task similarity between the two datasets. This regularizer is defined as
\vspace{-5pt}
\begin{equation}
    \mathcal{L}_{\text{sim}}(w_t) =  \left\| r_{ij} - d(g_{w_t}\left( \mathcal{D}_i \right),  g_{w_t}\left( \mathcal{D}_j \right)) \right\|_2^2,
    \label{eq:sim_fn}
\end{equation}
where $r_{ij}$ is a pre-computed similarity between the $i$-th and $j$-th datasets, and $d\left( g_{w_t}\left( \mathcal{D}_i \right),  g_{w_t}\left( \mathcal{D}_j \right) \right)$ is the cosine similarity between the two task representations $g_{w_t}\left( \mathcal{D}_i \right)$ and $g_{w_t}\left( \mathcal{D}_j \right)$. We pre-compute $r_{ij}$ as $AP@K$~\cite{zhu2004recall} with $k=10$. Intuitively, $r_{ij}$ is high when the top-$k$ configurations of the two datasets have a large number of entries in common. $\mathcal{L}_{\text{sim}}(w_t)$ thus helps $g_{w_t}(.)$ push an unseen dataset close to a ``similar'' seen dataset in the manifold, thereby improving hyperparameter recommendation for this new dataset. 

{\noindent \bf Reducing intra-task representation variance.} In order to optimize Eq.~\eqref{eq:loss_fn}, we leverage stochastic gradient descent with mini-batch size B. Consequently, this imposes the constraint that a dataset representation $\mathbf{t}_i$ computed from a batch sampled from a dataset $\mathcal{D}_i$ has to be representative of that dataset. In other words, a dataset representation $\mathbf{t}_i^a$ computed from a batch $a$ sampled from $\mathcal{D}_i$ has to be similar to a representation $\mathbf{t}_i^b$ computed from a batch $b$ of the same dataset. Therefore, our model has to ensure that the variance among the task representations computed from any batch of the same dataset has to be small. Inspired by domain adaptation techniques~\cite{ganin2014unsupervised, tzeng2017adversarial, hoffman2018cycada}, we devise an adversarial training component with the goal of keeping the dataset representations computed from batches ($\mathbf{t}_i^l$) close to the global representation of the dataset ($\mathbf{t}_i^G$). We compute the global representation of the dataset as follows $\mathbf{t}_i^G = \frac{1}{L}\sum_{l=1}^L \mathbf{t}_i^l$,
where the index $l$ runs through up to last $L$ sampled image batches of a dataset~(like a sliding window).
We use a discriminator $d_{w_d}(\cdot)$ to ensure that the batch-wise dataset representations $\mathbf{t}_i^l$ are close to the global representation $\mathbf{t}_i^G$. To penalize deviations, we formulate the following loss:
\vspace{-5pt}
\begin{equation}
\small
    \mathcal{L}_{\text{adv}}(w_t,w_d) = \mathbb{E} \left[\log\left(d_{w_d} \left( \mathbf{t}_i^G \right)\right)\right] + \mathbb{E} \left[\log \left(1 -  d_{w_d}\left( \mathbf{t}_i^l \right) \right)\right],    
\end{equation}
where $\mathbb{E}\left[\cdot\right]$ is the expectation operator. We chose to use an adversarial training component to ensure semantic consistency between batch-wise representations $\mathbf{t}_i^l$ and the global representation $\mathbf{t}_i^G$ as suggested by Hoffman~\etal~\cite{hoffman2018cycada}.

{\noindent \bf Overall objective function.} The overall task representation problem is thus the following
\begin{equation}
\small
    \min_{w_t,w_s,\theta} \max_{w_d} \quad \mathcal{L}_{\text{perf}}(w_t,w_s,\theta) + \alpha \mathcal{L}_{\text{sim}}(w_t) + \beta \mathcal{L}_{\text{adv}}(w_t,w_d)
    \label{eq:overall_loss_fn}
\end{equation}
where $\alpha$ and $\beta$ are loss coefficients. We solve this problem by alternating between optimizing feature extractors $g_{w_t}(\cdot)$, $h_{w_s}(\cdot)$ and discriminator $d_{w_d}(\cdot)$ until convergence. 

{\noindent \bf Implementation details of offline meta-learning phase.}
Algorithm~\ref{algo:offline} shows the training process for the offline meta-learning phase. The offline meta-learning phase requires two loops. The outer-most for-loop (steps 3 - 12) samples a meta-batch $\mathcal{C}_m$ for the $m$-th dataset $\mathcal{D}_m$ containing hyperparameter configurations and their performances. The inner-most for-loop (steps 6 - 11) samples image batches from $\mathcal{D}_m$ to update the parameters of the predictor and simultaneously aggregate the global representation $\mathbf{t}_m$ considering up to the $L$ last image batches. In addition to leveraging stochastic gradient descent as described above, sampling image batches to represent a dataset, compared to a single-point estimate~\cite{wong2018transfer}, helps the dataset~(task) to be modeled as a richer distribution in the dataset-configuration space by effectively acting as data augmentation.



\subsection{Online Recommendation Phase}
\vspace{-1mm}
Once the performance predictor of our HyperSTAR learns to effectively map a dataset-configuration pair to its corresponding performance score in the offline meta learning phase, we can use it for online recommendation on an unseen dataset $\mathcal{D}_{\text{new}}$ as shown in Algorithm~\ref{algo:online} and Figure~\ref{fig:overview}b.  HyperSTAR first extracts a task representation $\mathbf{t}_{\text{new}} = g_{w_t}(\mathcal{D}_{\text{new}}) $ for the new dataset and then along with a batch of previously-seen hyperparameter configuration encodings, feeds it into the offline-trained performance predictor $f_\theta(\cdot)$ to predict a sequence of performance scores corresponding to the sequence of configurations. Based on these performance scores, we can rank the configurations to prioritize which ones to evaluate.

{\noindent \bf  Task-Aware HPO.} This task-aware recommendation list generated by HyperSTAR can be used to warm-start and guide any of the existing HPO approaches. We prove this by proposing a task-aware variant of Hyperband~\cite{li2016hyperband}. In this variant of Hyperband, in each stage, we replace the random configuration sampling by evaluating the top $n$ configurations based on the recommendation list suggested by HyperSTAR. We experiment with a thresholded list of top $n$ configurations with Hyperband, but it can be hybridized (without much effort) to either mix a certain ratio of random configurations or sample configurations based on a probability defined over the ranked configuration list.

{\noindent \bf Implementation details of online phase.} Algorithm~\ref{algo:online} summarizes the online recommendation phase. The outer-most for-loop (steps 2 - 9) iterates over all the possible $H$ configurations. For each configuration, the inner-most loop (steps 4 - 7) samples $B$ batches and predicts the performance for each batch at step 6. At the end of this inner-most loop, we average all the performance predictions and use it as the performance estimate for the $n$-th configuration. Lastly, the algorithm ranks all the configurations based on their estimated performances and returns the ranking.

\begin{algorithm}[t]
\label{algo:offline}
\SetAlgoLined
\footnotesize{
\textbf{Input}  meta-dataset $\mathcal{T}$, $M$ datasets, hyperparameter batch size $O_{\text{hyper}}$, image batch size $N_{\text{img}}$, number of sampled image batches per dataset $B_{\text{img}}$, window size $L$ \\

\While{Not converge}{
\For{m=1 to M}{   
  Initialize global task embedding $\mathbf{t}^G_{m} = \mathbf{0} $\\ 
  Sample hyperparameter batch $\mathcal{C}_m$ from $\mathcal{T}$ for dataset $\mathcal{D}_m$\\
\For{i=1 to $B_{\text{img}}$}{  
 Sample an image batch $\mathcal{B}^i_m$ from $\mathcal{D}_m$ \\
   Update $\theta, w_t, w_s$ by minimizing Eq. \eqref{eq:overall_loss_fn} \\
 Compute $\mathbf{t}^G_{m}$ as mean of up to last $L$ image batches \\
    Update $w_d$ by maximizing Eq. \eqref{eq:overall_loss_fn};

}
}
 }
 }
 \caption{Offline Meta-learning Phase}
\end{algorithm}

\begin{algorithm}[t]
\label{algo:online}
\SetAlgoLined
\footnotesize{
\textbf{Input}  Unseen dataset $\mathcal{D}_{\text{new}}$, meta-dataset $\mathcal{T}$, batch sampling iterations $B$, number of hyperparameter configurations $H$  \\

\For{n=1 to H}{
  Get the $n$-th hyperparameter configuration $\mathbf{c}_n $ from $\mathcal{T}$\\
\For{i=1 to $B$}{  
 Randomly sample an image batch $\mathcal{B}^{i}_{\text{new}}$ from $\mathcal{D}_{\text{new}}$ \;
 $v_{n,i} = f_\theta \left(g_{w_t}\left( \mathcal{B}^{i}_{new} \right), h_{w_s}(\mathbf{c}_n)\right)$
}
 $v_{n} = \frac{1}{B} \sum^{B}_{i} {v_{n,i}}   $
}

Return ranked configurations $\mathbf{c}_1,\hdots, \mathbf{c}_H$ based on $v_1, \hdots, v_H$\\
}
 \caption{Online Recommendation Phase}

\end{algorithm}
 \vspace{-5pt}

\section{Experiments}
\vspace{-1mm}
\label{experiments}

\newcommand{\STAB}[1]{\begin{tabular}{@{}c@{}}#1\end{tabular}}

\begin{table*}[]
\scriptsize
\setlength\tabcolsep{2pt}
\centering
\caption{AP@10 comparison for SE-ResNeXt-50 for 10 public image classification datasets across different methods.}
\begin{tabular}{llcccccccccc|c}
\toprule
& Test Dataset         & BookCover30 & Caltech256 & DeepFashion & Food101 & MIT Indoor & IP102 (Pests) & Oxford-IIIT Pets & Places365 & SUN397 & Textures (DTD) & Average \\ \midrule
\multirow{5}{*}{\STAB{\rotatebox[origin=c]{90}{Baselines}}}
&  Feurer \etal\cite{feurer2015initializing} &      38.71       &  60.59           &     33.27        &     48.01   &         67.81     &       68.15        &        71.98          &     49.20      &   72.63    &      59.38        &    59.67   \\
& Feurer \etal\cite{feurer2015efficient}         &       37.13      &     49.55      &       28.67      &    49.60     &        43.27       &         50.71      &        54.50        &    54.78     &   54.78   &      51.97         &   42.49    \\
& Task-Agnostic  &      45.63       &   65.90       &   31.96       &      55.28  &        43.51                     &      63.23          &      53.90     &     31.96   &         45.58       &   60.07    & 49.70  \\ 
& Meta-data &      42.10       &  72.57           &     46.69        &     63.32    &          72.31     &       73.09        &        78.11          &     51.78      &   88.59     &       60.13         &     64.87    \\
& Global Mean         &       61.64      &     85.26       &       44.64      &    63.95     &        \textbf{79.41}       &         \textbf{89.05}      &      78.33            &     62.32      &    93.36    &       74.66         &    73.24     \\ \hline
\multirow{4}{*}{\STAB{\rotatebox[origin=c]{90}{Ablations}}}
& Batchwise Mean~(BM)  &      68.16       &    82.34        &     62.39        &     \textbf{70.98}    &        72.51                     &       84.94           &      \textbf{81.43}    &     88.05   &         93.74       &   \textbf{82.59}    & 78.71  \\
& BM + GAN &      64.02       &     83.83       &       87.63      &   67.27      &      76.45         &       87.49        &                 78.42  &   \textbf{93.41}        &  92.92      &       77.21         &    80.87     \\
& BM + Similarity        &     62.60        &    80.97        &      82.39       &   67.31      &       78.79        &       83.64                   &      79.52     &    90.37    &      \textbf{94.47}          &    81.63 & 80.17     \\
& BM + Similarity + GAN             &     \textbf{68.27}   &  \textbf{86.72}     &          \textbf{91.51} &  68.20     &    77.97      &    87.52      &    79.64         &     91.72      &    91.85    &      81.46          & \textbf{82.49} \\ \bottomrule      
\end{tabular}
\label{table:baseline_ablation_resnext}
\vspace{-10pt}
\end{table*}

\begin{table*}[]
\scriptsize
\setlength\tabcolsep{2pt}
\centering
\caption{AP@10 comparison for ShuffleNet-v2-x1 for 10 public image classification datasets across different methods.}
\begin{tabular}{llcccccccccc|c}
\toprule
& Test Dataset         &  BookCover30 & Caltech256 & DeepFashion & Food101 & MIT Indoor & IP102 (Pests) & Oxford-IIIT Pets & Places365 & SUN397 & Textures (DTD) & Average \\ \midrule
\multirow{5}{*}{\STAB{\rotatebox[origin=c]{90}{Baselines}}}
& Feurer \etal\cite{feurer2015initializing} &      14.30       &  3.41           &     6.75       &     12.57    &         9.11     &      14.21       &        2.11          &   25.81     &   18.45     &      59.38        &    11.72   \\
& Feurer \etal\cite{feurer2015efficient}         &      21.74     &    10.95     &      0.00     &    11.81     &        25.74       &         0.0      &         11.95         &   27.87   &   27.87   &       0.0        &   15.25    \\
& Task-Agnostic Baseline  &    0.00       &   2.11       &   15.74       &    44.69   &       3.11                    &     26.11          &      31.24   &    0.00  &       0.00     &   18.65    & 14.16 \\ 
& Meta-data &       33.27      &     15.35       &   31.97          &     33.44    &      35.68         &     54.73          &         25.46         &     40.48      &   37.89     &    34.88            &    34.32     \\
& Global Mean         &     35.30        &    16.45        &    17.70         &    47.41     &     34.87          &    49.87           &        24.98          &      58.28     &     43.79   &      40.71          &  36.94       \\ \hline
\multirow{4}{*}{\STAB{\rotatebox[origin=c]{90}{Ablations}}}
& Batchwise Mean~(BM) &      76.81      &     21.69       &     33.82        &    80.98     &     39.56          &  56.30           &      44.86            &     69.88      &    \textbf{50.49}    &     46.70           &   52.11      \\
& BM + GAN  &    69.71         &   \textbf{24.29}         &      34.34     &   84.88     &  46.93           &    54.96           &    53.84            &  62.89         &  43.98      &   \textbf{ 51.43}            &     52.72   \\
& BM + Similarity        &   \textbf{ 80.23}
&   20.51        &     35.28 &   \textbf{87.10}&    38.21
&     \textbf{ 57.73} &  \textbf{54.69}
&   71.78 &    44.78&       46.52&   53.68 \\
& BM + Similarity + GAN             &  76.92     &  21.51     &   \textbf{40.10}      &   84.60   &    \textbf{47.74 }   &   55.34      &   47.20        &    \textbf{75.25}  &  46.09    & 46.56          & \textbf{54.13} \\ \bottomrule      
\end{tabular}
\label{table:baseline_ablation_shufflenet}
\vspace{-10pt}
\end{table*}

This section presents a series of experiments designed to evaluate the performance predictor, the generated recommendation (see Sec.~\ref{sec:dataset_embeddings}), and the end-to-end HPO performance of HyperSTAR (see Sec.~\ref{sec:hpo_experiment}). 

{\noindent \bf Datasets.}~We evaluate HyperSTAR on 10 publicly available large-scale image classification datasets: BookCover30~\cite{iwana2016judging}, Caltech256~\cite{griffin2007caltech}, DeepFashion~\cite{liuLQWTcvpr16DeepFashion}, Food-101~\cite{bossard14}, MIT Indoor Scene Recognition~\cite{quattoni2009recognizing}, IP102 Insects Pests~\cite{wu2019ip102}, Oxford-IIIT Pets~\cite{parkhi12a}, Places365~\cite{zhou2017places}, SUN397~\cite{xiao2010sun} and Describable Texture Dataset~(DTD)~\cite{cimpoi14describing}. 


{\noindent \bf Architectures.} To ensure that our empirical study reflects the performance of our method on state-of-the-art network architectures, we choose SE-ResNeXt-50~\cite{hu2018squeeze}, a powerful and large architecture; and ShuffleNet-v2-x1~\cite{ma2018shufflenet}, a compact and efficient architecture. For both networks, we operate in a transfer-learning setting~\cite{donahue2014decaf} where we initialize the weights of the network from a model pre-trained on ImageNet~\cite{deng2009imagenet} and fine-tune certain layers of the network while minimizing the multi-class cross entropy loss. The hyperparameter space for SE-ResNeXt-50 consists of 40 configurations varying in learning rate, choice of optimizer, number of layers to fine tune, and data augmentation policy. As ShuffleNet-v2-x1 takes less time to train, we explore a larger search space of 108 configurations over the aforementioned hyperparameter dimensions.


{\noindent \bf Meta-dataset for training the performance predictor.} To construct the meta-dataset over the joint dataset-hyperparameter space, we train both SE-ResNeXt-50 and ShuffleNet-v2-x1 for every configuration in their respective hyperparameter space on each of the 10 datasets. This generates a set of 400 training samples for SE-ResNeXt-50 and 1,080 data samples for ShuffleNet-v2-x1. The meta-dataset thus contains triplets holding a one-hot encoding representing the hyperparameter configuration, training images of the dataset and corresponding Top-1 accuracy (to be used as performance score that HyperSTAR estimates).
We normalize these Top-1 accuracies using the mean and standard deviation computed for each dataset separately over the accuracy scores across the configuration space (Section~\ref{sec:offline_meta}). 

{\noindent \bf Evaluation metric.} We use Average Precision @ 10 ($AP@10$)~\cite{zhu2004recall} metric (as described in Section~\ref{sec:offline_meta}) to assess the ranking of configurations that HyperSTAR produces. This metric reflects the quantity and relative ranking of the relevant configurations predicted by HyperSTAR. We first build a ground-truth hyperparameter recommendation list based on decreasing order of actual Top-1 accuracies. We then compute $AP@10$ by comparing this list with the predicted recommendation list generated based on the decreasing order of the predicted accuracies from HyperSTAR.

\begin{figure*}[t]
    \centering
    \includegraphics[width=\textwidth]{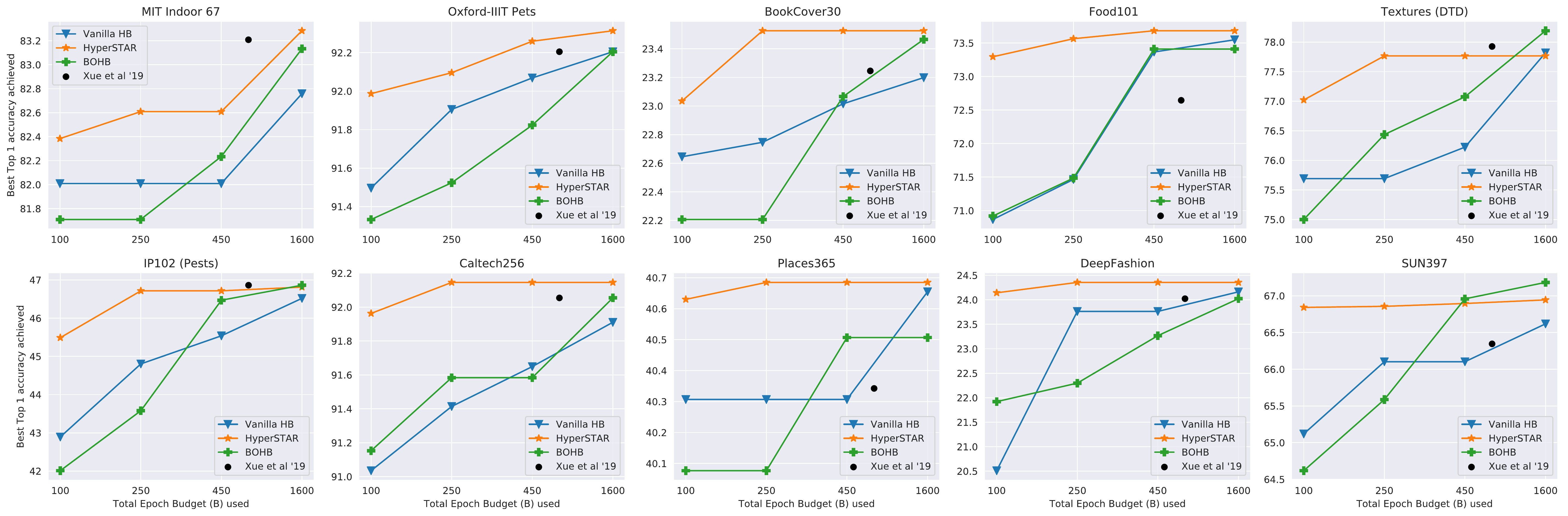}
    \caption{End-to-End performance comparison of task-aware HyperSTAR based Hyperband with existing methods for SE-ResNeXt-50. HyperSTAR outperforms other methods when performing on low epoch budgets~(100, 250, 450).}
    \label{fig:hb_seresnext50}
    \vspace{-10pt}
\end{figure*}

\begin{figure*}[t]
    \centering
    \includegraphics[width=\textwidth]{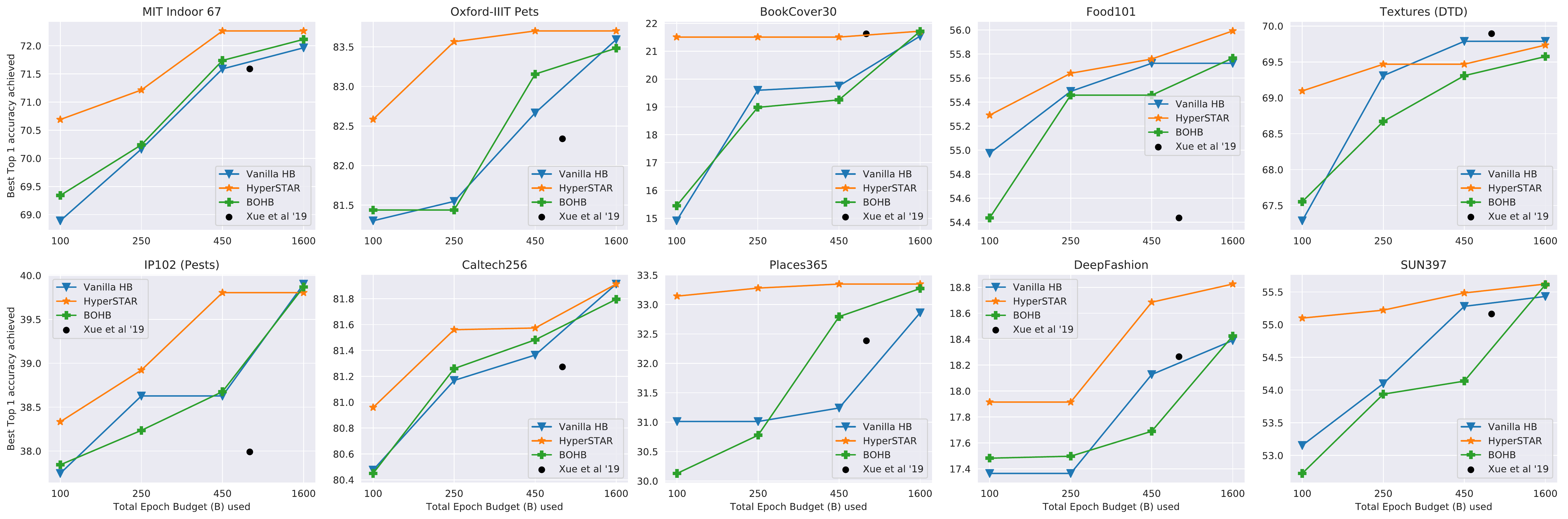}
    \caption{End-to-End performance comparison of task-aware HyperSTAR based Hyperband with existing methods for ShuffleNet-v2-x1. HyperSTAR outperforms other methods when performing on low epoch budgets~(100, 250, 450).}
    \label{fig:hb_shufflenet}
    \vspace{-10pt}
\end{figure*}

\vspace{-1mm}
\subsection{Performance Predictions and Rankings}
\label{sec:dataset_embeddings}
\vspace{-1mm}
We present a quantitative study comparing the task representation, regularization functions and the performance predictor introduced by HyperSTAR with existing methods. 

{\noindent \bf Task Representation Comparison.} For this comparison, we use meta-data based task representation as the first baseline. The representation is a subset of statistics used in previous methods \cite{feurer2015initializing,bardenet2013collaborative} which are applicable to our vision datasets (such as number of images, classes and images per class in the dataset). As the second baseline, we consider global mean features based task representation. The global mean is computed by taking an average of the deep visual features obtained from the penultimate layer of ResNet-50 pretrained on ImageNet~\cite{wong2018transfer} over all training images of a dataset. In comparison, our task representation is a batchwise mean (BM) taken as mean over end-to-end learned features over a batch of $N_{\text{img}}=64$ training images. We take $B_{\text{img}}=10$ of these batches and take an average to obtain the task representation. For training, the size of the hyperparameter batch is $O_{\text{hyper}}=10$. For each setting, we train our performance predictor and compute $AP@10$ averaged over 10 trials. We can observe from Tables~\ref{table:baseline_ablation_resnext} and~\ref{table:baseline_ablation_shufflenet} that our end-to-end learned task representation (BM) outperforms meta-data-based and global-mean-based task representations by  $17.62\%$ and $9.25\%$, respectively, for SE-ResNeXt-50. The performance gains are similar for ShuffleNet-v2-x1 (see Table~\ref{table:baseline_ablation_shufflenet}). This suggests that learning end-to-end visually inspired task representation helps HyperSTAR to recommend better task-aware configurations. It further suggests that representing the dataset as a distribution over a large number of randomly sampled batches is better than representing it as a point estimate using global mean.

{\noindent \bf Regularization Ablation Study.} We perform an internal ablation study comparing AP@10 achieved when using batchwise mean (BM) in HyperSTAR with and without imposing similarity and adversarial based regularization. We can observe from Tables~\ref{table:baseline_ablation_resnext} and~\ref{table:baseline_ablation_shufflenet} that imposing the regularizations improves the AP@10 for 6 out of 10 datasets for SE-ResNeXt-50 and 9 out of 10 dataset for ShuffleNet-v2-x1. This suggests that, on expectation, imposing regularization allows the task representations to learn meaningful features over the joint dataset-configuration space. Although there is a time cost associated with introducing regularizations, they provide an added dimension for the user to explore and further improve the configuration recommendation compared to the plain batch-wise mean setting.

{\noindent \bf Performance Predictor Comparison.} We compare HyperSTAR and existing meta-learning based warm-starting HPO methods. We first compare HyperSTAR with Feurer~\etal\cite{feurer2015initializing} that uses random forest regression over a joint vector of meta-data and one-hot encoding of hyperparameters to predict the corresponding Top-1 accuracy and use that to build a recommendation list of configurations. We also compare HyperSTAR with Feurer~\etal\cite{feurer2015efficient} that finds the most similar training dataset for a given test dataset with respect to meta-data features and use the ground truth list of recommended configurations for the training dataset as the prediction for test dataset. We further set up a task-agnostic baseline to compare the effectiveness of our task-aware recommendations. For this baseline, we disregard the characteristics of the test dataset and build a predicted list of recommended configurations by sorting the configurations in decreasing order of their average Top-1 accuracy over the training datasets. From Table ~\ref{table:baseline_ablation_resnext} and ~\ref{table:baseline_ablation_shufflenet}, we can observe that HyperSTAR surpasses each of the baselines with average AP@10 margin of at least 25\% for SE-ResNeXt-50 and 37\% for ShuffleNet-v2-x1. We also observe from the Tables that similarity based approaches (task-agnostic and Feurer~\etal\cite{feurer2015efficient}) have a higher variance in performance across datasets compared to task-representation-based approaches (HyperSTAR and Feurer~\etal\cite{feurer2015initializing}). 

\vspace{-1mm}
\subsection{Warm-Starting with Recommendations}
\vspace{-1mm}
We test the configurations recommended by HyperSTAR and other baseline methods by evaluating their ranking order. We plot a curve showing the best Top-1 accuracy achieved after $k$ hyperparameter configurations for $k = 1 \ldots H$ as shown in Figure~\ref{fig:baseline_comparison}a. We can observe that using the recommendation from HyperSTAR achieves the same performance in just 50\% of evaluated configurations as needed by baseline recommendations. This suggests that compared to other baseline methods and task representations, raw-pixel based end-to-end learned task representations of HyperSTAR are more informative for prioritizing hyperparameter configurations. HyperSTAR takes $422ms$ on an Nvidia 1080Ti to generate configuration recommendations which is negligible compared to multiple GPU hours required to evaluate even a single configuration.

\subsection{Task-Aware Hyperband}
\begin{figure}[t]
    \centering
    \includegraphics[width=\columnwidth]{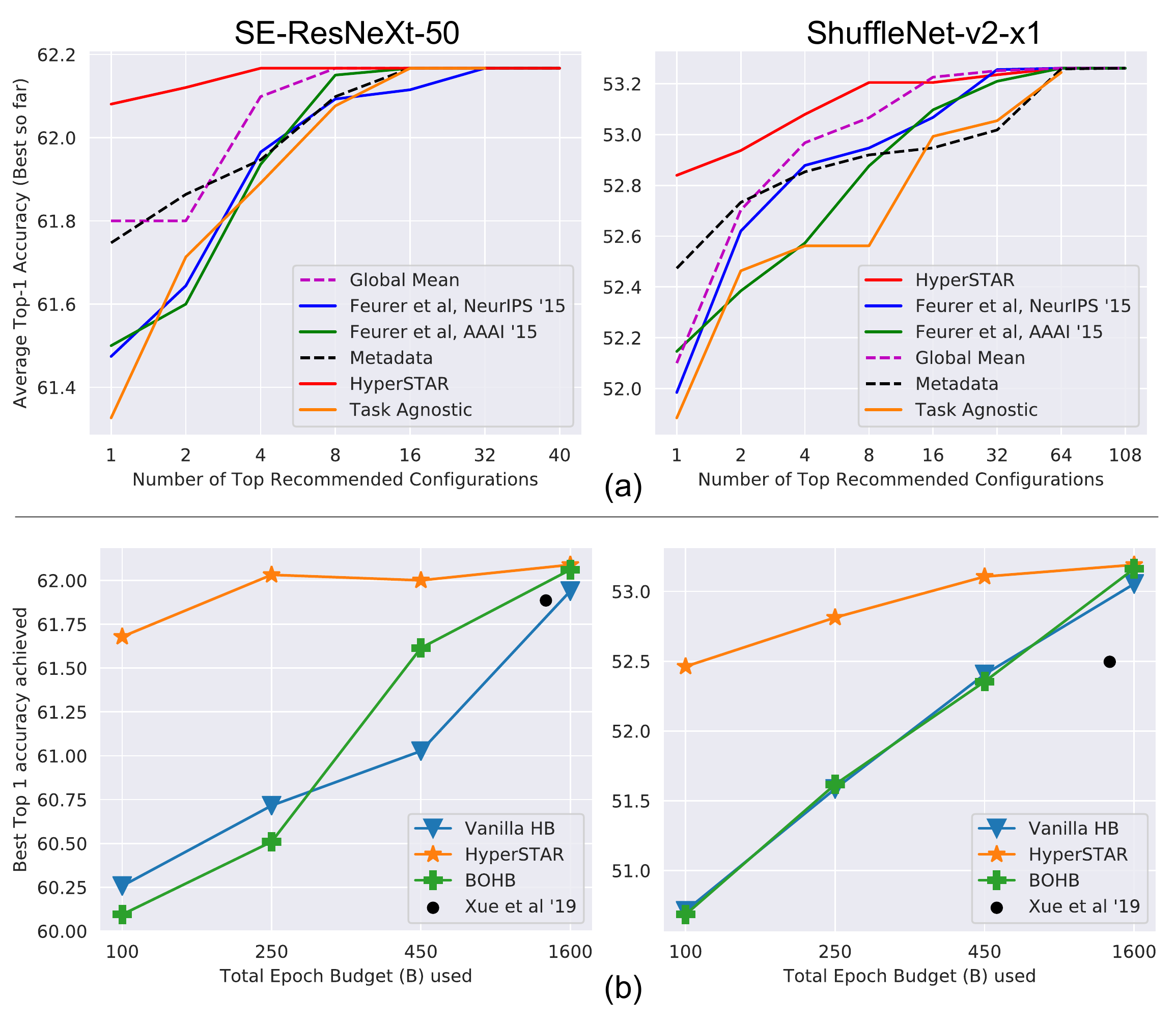}
    \caption{(a) Comparison of evaluating configurations recommended by HyperSTAR with baseline methods in ranked order. Compared to baselines, HyperSTAR achieves the best performance by evaluating $50\%$ less configurations. (b) Comparison of warm-starting Hyperband (HB) with HyperSTAR vs. baseline approaches across different epoch budgets. HyperSTAR achieves optimal accuracy in $25\%$ of the budget required by other methods. It also achieves $1.5\%$ higher best Top-1 accuracy on average for the smallest budget setting on both network architectures.}
    \label{fig:baseline_comparison}
\vspace{-4mm}
\end{figure}

\vspace{-1mm}
\label{sec:hpo_experiment}
We warm-start Hyperband (HB)~\cite{li2016hyperband}  using the task-aware hyperparameter recommendation from HyperSTAR and compare it with the vanilla Hyperband~\cite{li2016hyperband} and BOHB~\cite{falkner18abohb}. We design the experiment to demonstrate a common scenario where the time available to search for the optimal hyperparameters for an unseen dataset is limited. We run all the methods for different amounts of total budget. The budget is defined in terms of epochs to keep the evaluation time consistent across different hyperparameter configurations, datasets and architectures. The maximum number of epochs for any given configuration is $R=100$. We consider a budget of 1600 epochs ($\eta=3$, large-budget setting) and smaller budgets of 450, 200 and 100 epochs~($\eta=2$, low-budget settings). Figs.~\ref{fig:hb_seresnext50} and \ref{fig:hb_shufflenet} show the best Top-1 accuracy achieved by the different methods for different budgets for all 10 test datasets for SE-ResNeXt-50 and Shufflent-v2-x1, respectively. Fig.~\ref{fig:baseline_comparison}b further shows the average over all the test datasets for the two network architectures. We observe that HyperSTAR outperforms vanilla HB and BOHB in the low-budget settings for all datasets achieving around $1.5\%$ higher best Top-1 accuracy on average for the smallest budget setting on both network architectures. In fact, HyperSTAR achieves the optimal accuracy in just $25\%$ of the budget required by the other two methods. This happens because the initial set of hyperparameters suggested by both vanilla HB and BOHB do not follow any prior and are chosen randomly, \ie, they are task agnostic.

The difference in the Top-1 accuracy achieved by all three methods gradually diminish with increasing time budget and eventually becomes negligible for the largest budget setting. The accuracy is also at par with the best Top-1 accuracy achievable for the given hyperparameter space. This happens for vanilla HB and BOHB because over time, they explore the hyperparameter space sufficiently enough to be able to discover the best possible configuration. Although HyperSTAR-based Hyperband has been designed to improve HPO efficiency for low-budget setting, being able to achieve the best possible performance suggests that it is also sound for large-budget setting. Given sufficient budget, our method can  achieve at par (if not better) performance compared to other HPO methods. Our plots also show BOHB being comparable to vanilla HB for low-budget setting while being better than vanilla HB in the large-budget setting. This is because of the Bayesian sampling prior of BOHB that gets better than random sampling over time, thus helping BOHB outperform vanilla HB.

We also compare our task-aware Hyperband with Tr-AutoML~\cite{xue2019transferable}. For a fair comparison, we considered the time Tr-AutoML spends to group the 9 training datasets as part of offline training and exclude it from the time comparison. We randomly choose 10 configurations to group the datasets and evaluate on the test dataset for finding the most similar training dataset. We consider the more time efficient scenario of Tr-AutoML where we do not run Hyperband and compute the Top-1 accuracy achieved over the unseen dataset using the best configuration for the most similar training dataset. As shown in Figures~\ref{fig:hb_seresnext50}, \ref{fig:hb_shufflenet} and \ref{fig:baseline_comparison}b, since the total evaluation time comprises of running on the benchmark 10 configurations and then finally on the best found configuration, the Top-1 accuracy is reported as achieved after 1100 epochs. From the figures, we can observe that, on expectation, our task-aware HB is able to achieve the same performance in as little as $10$ times less budget. This reinforces that learning a dataset embedding from raw pixels significantly reduces the time required to predict the optimal hyperparameters compared to Tr-AutoML.

\vspace{-3mm}
\section{Conclusion}
\vspace{-1mm}


We present HyperSTAR, the first efficient task-aware warm-start algorithm for hyperparameter optimization (HPO) for vision datasets. It operates by learning an end-to-end task representation and a performance predictor directly over raw images to produce a ranking of hyperparameter configurations. This ranking is useful to accelerate HPO algorithms such as Hyperband. Our experiments on 10 real-world image classification datasets show that HyperSTAR achieves the optimal performance in half the number of evaluated hyperparameter configurations compared to state-of-the-art warm-start methods. Our experiments also show that HypterSTAR combined with Hyperband achieves an optimal performance in 25\% of the budget of other HB variants. HyperSTAR is especially helpful in performing HPO without requiring large computational time budgets. 


%



\vspace{-3mm}
\section{Acknowledgements}
\vspace{-1mm}
Special thanks to Microsoft Custom Vision team for their valuable feedback and support.

{\small
\bibliographystyle{ieee_fullname}
\bibliography{references}
}

\end{document}